%% file: conference_101719.tex
\documentclass[conference]{IEEEtran}
\IEEEoverridecommandlockouts
\usepackage{cite}
\usepackage{amsmath,amssymb,amsfonts}
\usepackage{algorithmic}
\usepackage{graphicx}
\usepackage{textcomp}
\usepackage{xcolor}
\usepackage{lipsum}
\usepackage{hyperref}
\hypersetup{
    colorlinks,
    linkcolor={black},
    citecolor={black},
    urlcolor={blue!80!blue}
}

\def\BibTeX{{\rm B\kern-.05em{\sc i\kern-.025em b}\kern-.08em
    T\kern-.1667em\lower.7ex\hbox{E}\kern-.125emX}}
\begin{document}

\title{Solarcast-ML: Per Node GraphCast Extension for Solar Energy Production
}

\author{\IEEEauthorblockN{1\textsuperscript{st} Cale Colony}
\IEEEauthorblockA{\textit{Robotics Department} \\
\textit{University of Michigan}\\
Ann Arbor, United States of America \\
ccolony@umich.edu}
\and
\IEEEauthorblockN{2\textsuperscript{nd} Razan Andigani}
\IEEEauthorblockA{\textit{Robotics Department} \\
\textit{University of Michigan}\\
Ann Arbor, United States of America \\
andigani@umich.edu}

}

\maketitle

\begin{abstract}
This project presents an extension to the GraphCast model, a state-of-the-art graph neural network (GNN) for global weather forecasting, by integrating solar energy production forecasting capabilities. The proposed approach leverages the weather forecasts generated by GraphCast and trains a neural network model to predict the ratio of actual solar output to potential solar output based on various weather conditions. The model architecture consists of an input layer corresponding to weather features (temperature, humidity, dew point, wind speed, rain, barometric pressure, and altitude), two hidden layers with ReLU activations, and an output layer predicting solar radiation. The model is trained using a mean absolute error loss function and Adam optimizer. The results demonstrate the model's effectiveness in accurately predicting solar radiation, with its convergence behavior, decreasing training loss, and accurate prediction of solar radiation patterns suggesting successful learning of the underlying relationships between weather conditions and solar radiation. The integration of solar energy production forecasting with GraphCast offers valuable insights for the renewable energy sector, enabling better planning and decision-making based on expected solar energy production. Future work could explore further model refinements, incorporation of additional weather variables, and extension to other renewable energy sources.

The project page is available at: \href{http://solarcast-ml.com}{http://solarcast-ml.com}.
\end{abstract}

\section{Introduction}
Weather has been a defining characteristic and focus of the human condition even before the advent of agriculture.  It is a truly universal phenomenon, with every human, animal, and plant impacted by the weather on a daily basis, a fundamental aspect of the environment in which we live.

In the modern era, governments spend billions of dollars attempting to predict the weather.  While we are separated from our environment more and more by air conditioned offices and heated homes, the weather still yields profound impacts on how we live our lives.

To date, most weather forecasting completed at an organizational scale uses numerical models which attempt to simulate the weather using a set of equations describing fluid behaviors at a planetary scale.  These models require extremely large amounts of computational resources, necessitating execution on fixed intervals.  For example, in the United States, the National Weather Service recently increased their computation capacity 10-fold\cite{kravets-2015}.

Weather being a dynamical system, is a ripe target for machine learning tasks.  The landmark study this work built a novel model based on a GNN called GraphCast.  GraphCast's primary output is the predicted state of the global weather in 6 hours given the inputs of the current weather, and the weather 6 hours previous.  This output can be used auto-regressively to predict the weather a further 6 hours in the future. This process can be iterated to continue forecasting weather reliably up to 10 days into the future.

One of the critical sectors that stands to benefit significantly from advancements in weather forecasting is renewable energy, particularly solar energy production. The ability to accurately predict solar radiation and energy output is crucial for optimal resource allocation, grid management, and energy storage strategies. However, solar energy production is highly dependent on various weather conditions, such as cloud cover, humidity, and atmospheric aerosols, making it a challenging task to forecast accurately.

This project aims to address this challenge by integrating solar energy production forecasting capabilities with the state-of-the-art GraphCast model, leveraging its accurate weather forecasts. By training a neural network model to predict the ratio of actual solar output to potential solar output based on various weather conditions, the proposed approach offers valuable insights for the renewable energy sector, enabling better planning and decision-making based on expected solar energy production

\section{Related Work}
    \input{related_work}

\section{Algorithmic Extension}
    \input{algorithmic_extension}

\section{Experiments and Results}

\subsection{Data Collection}
    \input{data_collection}

\subsection{Experimental Setup}
    \input{experiment_design}

\subsection{Results}
    \input{results}

\section{Conclusions}
    \input{Conclusion}

\bibliographystyle{IEEEtran}
\bibliography{biblio}

\end{document}

%% file: related_work.tex
Machine learning (ML) has emerged as a promising approach to weather forecasting, complementing traditional numerical weather prediction methods. ML models can can learn complex patterns and relationships from vast amounts of historical weather data, enabling them to make accurate predictions without explicitly modeling the underlying physical process \cite{cervone-2017}.

One of the early applications of ML in weather forecasting involved using artificial neural networks (ANNs) to predict surface temperature and precipitation \cite{maren-1990}. These models demonstrated the potential of ML to capture nonlinear relationship between weather variables and improve forecast accuracy compared to traditional statistical methods.

In recent years, deep learning (DL) techniques have gained popularity in weather forecasting due to their ability to learn hierarchical representations from raw data \cite{10.1145/2783258.2783275}. Convolutional Neural Networks (CNNs) have been employed to predict short-term precipitation \cite{shi2015convolutional}, while Recurrrent Neural Networks (RNNs) have been used to model the temporal dependencies in weather data \cite{shi2017deep}. These DL models have shown promising results in capturing spatial and temporal patterns, outperforming traditional ML approaches \cite{reichstein-2019}.

Graph Neural Networks (GNNs) have also been explored for weather forecasting, as they can effectively model the complex spatial dependencies between weather variables \cite{MA2023119580}. GNNs have been applied to predict various weather phenomena, such as tropical cyclones and extreme weather events \cite{gao-2018}, demonstrating their ability to capture the intricate relationship in weather systems. 

Hybrid approaches that combine ML with physical models have also been proposed to leverage the strengths of both methods. For example, \cite{bremnes} integrated a deep learning model with a numerical weather prediction system to improve the accuracy of short-term forecasts. Similarly, \cite{rangaraj-2024} used ML to bias-correct and downscale the outputs of a physical model, improving the spatial resolution and reducing systematic errors in the forecasts.

Despite the advancements in ML-based weather forecasting, challenges remain in terms of interpretability, generalization, and computational efficiency \cite{mcgovern-2017}. Efforts have been made to address these issues, such as using attention mechanisms to improve interpretability \cite{kalander2020spatiotemporal} and transfer learning to enhance generalization \cite{kumar2023precipitation}. Additionally, techniques like model compression and distributed training have been explored to reduce the computational burden of training large-scale ML models for weather forecasting \cite{TANG2023127864}.

GraphCast \cite{lam2022graphcast} represents a significant milestone in ML-based weather forecasting, demonstrating the potential of GNNs to learn skillful medium-range global weather forecasts. The GraphCast model employs a novel architecture that combines GNNs with multi-resolution icosahedral meshes, enabling it to capture the complex spatial and temporal dependencies in weather data effectively.

The GraphCast architecture consists of two main components: a GNN encoder and a GNN decoder. The encoder takes as input the current state of the weather and the state six hours prior, preprocessed into a graph representation using an icosahedral mesh. The encoder then processes this graph data through a series of graph convolutional and attention layers, learning a latent representation of the weather dynamics. The learned latent representation is then passed to the decoder, which is another GNN architecture responsible for generating the predicted state of the weather six hours into the future. The decoder operates in an auto-regressive manner, iteratively predicting the next time step based on the previous predictions and the learned latent representation from the encoder.

One of the key innovations of GraphCast is its use of multi-resolution icosahedral meshes to represent the weather data as graphs. This approach allows for efficient modeling of global weather patterns while maintaining high resolution in areas of interest. The mesh is adaptively refined or coarsened based on the spatial variability of the weather data, enabling GraphCast to capture both large-scale and localized weather phenomena effectively.

Our work builds upon the foundation laid by GraphCast and aims to extend its capabilities by integrating solar energy production forecasting. By leveraging the weather forecasts generated by GraphCast, we develop a model that predicts the actual solar output from current weather conditions. This extension has the potential to provide valuable insights for the renewable energy sector, enabling better planning and decision-making based on expected solar energy production.

Our approach differs from previous works in several aspects. First, we focus specifically on the relationship between weather conditions and solar energy production, whereas most prior studies have concentrated on general weather forecasting. Second, we leverage the state-of-the-art GraphCast model as the backbone for our extension, ensuring that we build upon the most advanced and accurate weather forecasting techniques available. Finally, our model outputs a single value the actual to potential solar output, providing a concise and interpretable metric for solar energy stakeholders.

%% file: algorithmic_extension.tex
Our algorithmic extension is based on a dataset that was generated for the purposes of this project.  Our goal was to gather data relating to the solar output as able to be gathered by solar panel.  There are several factors involved the amount of solar radiation at the surface.  Foremost, the amount of radiation from the sun varies over time and has a profound impact on weather and climate. Total solar irradiance (TSI) is the total amount of radiation at the top of the atmosphere.  Beyond that, the radiation at the top of the atmosphere can be attenuated by different atmospheric effects such as clouds, humidity, and dust.

To capture the relationship between weather conditions and solar radiation, the proposed approach aims to predict the actual solar output as a function of weather conditions. This output provides a concise and interpretable metric for solar energy stakeholders, enabling them to assess the performance of solar energy systems and make informed decisions regarding resource allocation and energy management.

Our proposed extension suffered from a data mismatch from the original training data supplied to train GraphCast and the data collected to implement this project.  Because we wanted to train our model to run behind GraphCast, we chose our inputs to be a subset of the union of GraphCast outputs and the available data points from our data collection.  Our first attempt at training the model was fruitful with the following data points: temperature, humidity, dew point, wind speed, precipitation event, barometer, and solar altitude ratio (sun position against sun position at solar noon).  Additionally, a timestamp was collected for organization purposes and a pyranometer measurement was collected to serve as a ground truth for training.


%% file: data_collection.tex
Our goal for this project is to be able to predict the amount of solar energy incident at the surface of the earth from the output of the Graphcast prediction model. The first step was to establish ground truth values and verify that they are available for training our extension model.  Our collection apparatus in-situ was an Ambient Weather WS-2902 weather sensing array.  The unit has sensors for temperature, humidity, wind speed, solar irradiance, UV levels, and wind direction.  The unit uses a 915MHz back haul to get the environmental information to the reporting device.  

\begin{figure}[t!]
\centering
\includegraphics[width=0.75\columnwidth]{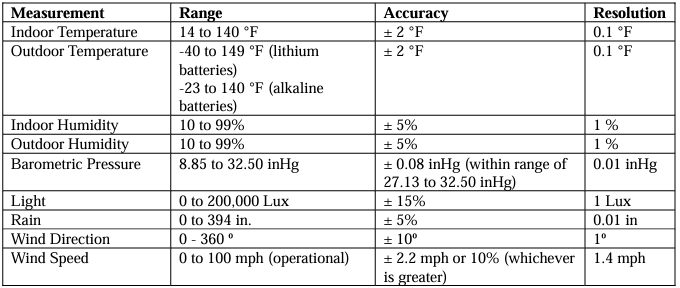}
\caption{Image detailing specifics of weather sensor}
\label{fig:teaser}
\end{figure}

The reporting device is responsible for collecting and collating the information from the different sensors located on premises. The reporting device used for this experiment is the Ambient Weather Observer IP module.  The Observer IP module allows for the transmission of data collected from the 915Mhz sensor band to the internet via wifi or ethernet.  The Observer IP module is by default setup to send data to the Ambient Weather network for display on user devices.  For this project, the reporting services were customized to send a POST request to an on-site virtual machine for data scrubbing and commitment to a custom data repository setup on Amazon Web Services (AWS) S3 storage service.  The unit was set to send the collated data every 16 seconds.

The POST requests from the ObserverIP module were send to a virtual server hosted on an ESXi host on premises.  The server was running Node-RED as the actual service responding to the POST requests.  The Node-RED service has the advantage of conveniently processing requests in JSON, which simplifies data collation and injection into AWS S3. Node-RED is built on Node.JS, hence, is message-based. Each message is broken into a message container and message payload. 

Each POST request received by the ObserverIP module starts a message chain that first triggers a sun-position node bound to the latitude and longitude of the weather station.  The output payload of this node is joined with the payload of the original weather reporting message payload and passed to the file-node which writes the completed message to a temporary file.  The lastUpdateStr is then extracted from the message payload and used to set a msg.filename variable used by the S3 upload node.  A failure of design was the inclusion of colons in the lastUpdateStr, which is allowed on the S3 system, but causes sync failures when pulling the data out of the S3 bucket back to Windows machines.  After the temporary file is created, the file is transmitted to the S3 bucket s3://deeprob-weatherdata/.  This was the final repository of the experimental data and data collection is planned to continue indefinitely.  While there was no accommodation for buffering information locally on-site, local infrastructure has redundant ISP access, which resulted minimal data loss caused by a lack of communication with AWS S3.

%% file: experiment_design.tex
The experimental setup consists of training a neural network model to predict solar radiation based on weather conditions data collected by the Ambient Weather WS-2902 weather sensing array. The dataset contains nine features: timestamp, solar radiation, temperature (F$^{\circ}$), humidity, dew point (F$^{\circ}$), wind speed (mph), rain (in), barometric pressure (in), and solar altitude percent.

Using the ground truth provided by the implemented data collection system, we designed a model that took the weather data from both current conditions and data 6 hours in the past for the location located above the ground truth weather station.  GraphCast will use these chronological data points to generate a global forecast, but our extension model is only concerned with the data above the ground truth weather station. As depicted in Figure 2, the weather sensing array is installed at a specific location, with measurements indicating a distance of 481 feet and a defined path from the array to the data collection point.

Graphcast was originally trained using 32 TPUv4 nodes over the course of three weeks. As such, the retraining of the original model was outside of the computation ability of our team.  The execution alone of the full GraphCast model requires ~25GB of GPU ram, placing it above the highest-grade consumer model available as the NVIDIA GeForce RTX4090. Due to these computational constraints, we focused on developing the extension model to predict solar radiation using the weather conditions data. 

Data was collected from the Copernicus Climate Data Store (CDS), which provides access to various climate datasets. Requests were submitted to the offline database, with each taking approximately 30 minutes to process. Due to the limitations of the CDS system, the dataset could only be downloaded in 4-day increments, with each day requiring approximately 30kb of storage. To ensure compatibility with the original GraphCast model, the dataset was zeroed on the area corresponding to the ground truth location, and the grid layout was matched to the GraphCast. The specific latitude and longitude coordinates for the datapoint were 42.56000137, -83.63999939, which were located 481 feet from the ground truth weather station. The collected data was validated against commercially available almanacs using the Panoply tool provided by NASA, ensuring the accuracy and reliability of the dataset.

\begin{figure}[t!]
    \centering
    \includegraphics[width=0.75\columnwidth]{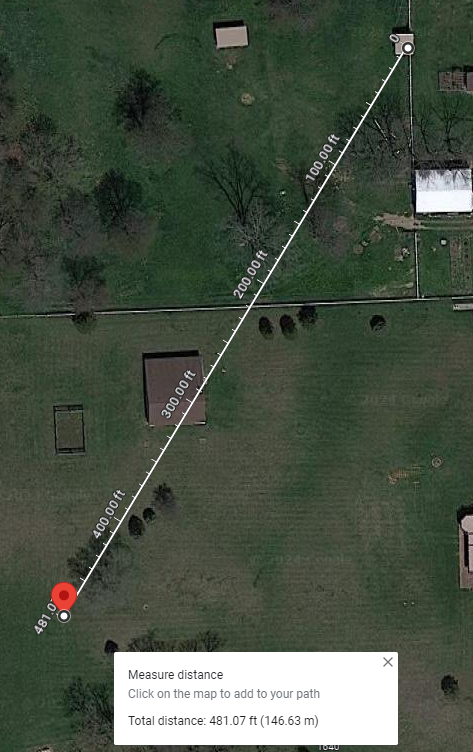}
    \caption{Aerial view of the weather sensing array location, along with measurements indicating the distance and path from the array to the GraphCast prediction point provided in the ERA5 dataset.}
    \label{fig:teaser}
\end{figure}

The dataset collected on-site for this project was preprocessed and transformed into a suitable format for training the neural network model. A custom WeatherDataset class was implemented, inheriting from the PyTorch Dataset class. This custom dataset class takes the preprocessed data and returns the weather conditions (temperature, humidity, dew point, wind speed, rain, barometric pressure, and altitude percent) as input features and the corresponding solar radiation as the target variable. By excluding the timestamp and solar radiation from the input features, the model learns to predict solar radiation based solely on the weather conditions.

The neural network model was constructed using the PyTorch nn.Module class, allowing for flexible and customizable architecture design. The model consists of a sequence of linear layers with Rectified Linear Unit (ReLU) activations, which introduce non-linearity and enable the model to learn complex patterns in the data. The input layer has 7 neurons, corresponding to the 7 input features (weather conditions), followed by two hidden layers, each containing 32 neurons. The output layer consists of a single neuron, representing the predicted solar radiation value. To initialize the weights of the linear layers, the Kaiming normal initialization method was employed, which has been shown to improve the convergence and stability of deep neural networks. The biases of the linear layers were initialized to zero.

Other hyperparameters were also tried.  The tested number of hidden layers varied from one to three.  The number of neurons in the hidden layer were tested from 7-256 along powers of two.  An extra ReLU layer after the final Linear was also checked, the theory being that solar output can never be below zero.  Additionally, a full suite of tests were ran checking the convergence behavior of the SGD.  Interestingly, SGD convergence started off better, but would eventually spiral out to large losses and get "frozen" with further epochs no longer changing the loss.

The model was trained using the Mean Absolute Error (MAE) as the loss function, which measures the average absolute difference between the predicted and actual solar radiation values. The choice of MAE as the loss function is appropriate for this regression task, as it provides a clear and interpretable measure of the model's prediction accuracy. Two optimization algorithms were compared - SGD with a learning rate of 0.001 and batch size 128, and the Adam algorithm. The model was trained for 100 epochs for evaluation of hyperparameters, then trained for 1000 epochs on final hyperparameter choices for a final loss value.

%% file: results.tex
Figure 3 shows the convergence of the model's loss during training using the Stochastic Gradient Descent (SGD) optimizer over 100 epochs. The plot displays the average loss for each epoch, which fluctuates but generally decreases, indicating that the model is gradually improving its ability to predict solar radiation accurately.  The model did not however, show classic forms of convergence over time regardless of the hyperparameters chosen.

\begin{figure}[t!]
    \centering
    \includegraphics[width=0.75\columnwidth]{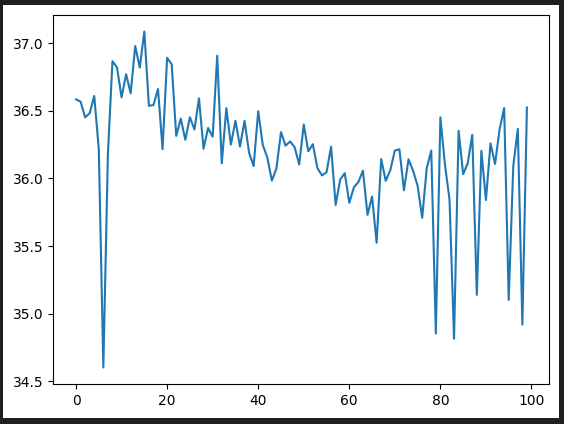}
    \caption{Convergence of the model using SGD optimizer over the course of 100 epochs.  Y-axis is watts per square meter mean square training loss.}
    \label{fig:teaser}
\end{figure}

Figure 4 depicts the convergence of the model's loss when trained using the Adam optimizer, an adaptive learning rate optimization algorithm. Similar to the SGD case, the loss decreases rapidly in the initial epochs and then flattens out, suggesting that the model has converged and learned to map the input weather conditions to the target solar radiation values effectively.  Training with the Adam optimizer more clearly showed convergence, and ultimately was used in the final training of the model.  

\begin{figure}[t!]
    \centering
    \includegraphics[width=0.75\columnwidth]{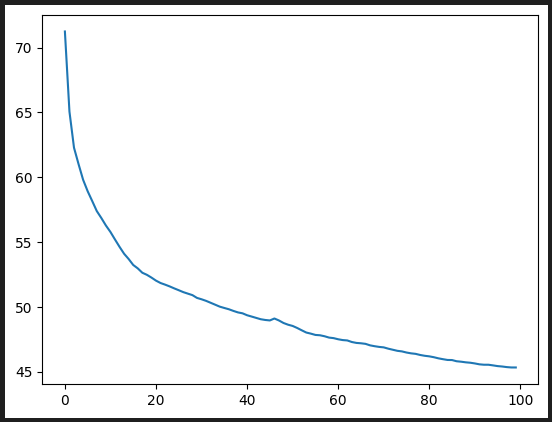}
    \caption{Convergence of the model using Adam optimizer over the course of 100 epochs.  Y-axis is watts per square meter mean square training loss.}
    \label{fig:teaser}
\end{figure}

The final convergence of the trained model is presented in Figure 4, which shows a plot of the predicted solar radiation values against the data points. The predicted values closely align with the ground truth solar radiation measurements, indicating the effectiveness of the proposed approach in integrating solar energy production forecasting with the GraphCast weather forecasting model.

\begin{figure}[t!]
    \centering
    \includegraphics[width=0.75\columnwidth]{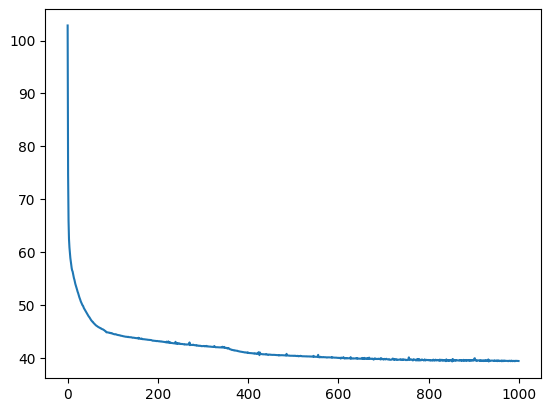}
    \caption{Final convergence of the trained model over 1000 epochs, showing predicted solar radiation values.  Y-axis is watts per square meter mean square training loss.}
    \label{fig:teaser}
\end{figure}

Overall, the results demonstrate that the proposed neural network model, trained using the Adam optimizer, is capable of effectively predicting solar radiation based on various weather condition inputs. The model's convergence behavior, decreasing training loss, and accurate prediction of solar radiation patterns suggest that it has successfully learned the underlying relationships between weather conditions and solar radiation.

%% file: Conclusion.tex
This work presented an extension to the GraphCast model, a state-of-the-art graph neural network for global weather forecasting, by integrating solar energy production forecasting capabilities. The proposed approach aimed to leverage the weather forecasts generated by GraphCast and train a neural network model to predict the actual solar output based on Graphcast outputs.

The neural network model was designed with an input layer corresponding to weather condition features (temperature, humidity, dew point, wind speed, event precipitation, barometric pressure, and solar altitude), followed by two hidden layers with ReLU activations and an output layer predicting the solar radiation value. The model was trained using a mean absolute error loss function and optimized with the Adam optimizer.

The results demonstrated the effectiveness of the proposed approach in accurately predicting solar radiation based on weather conditions. The model's convergence behavior, decreasing training loss, and accurate prediction of solar radiation patterns suggest that it has successfully learned the underlying relationships between weather conditions and solar radiation.

The integration of solar energy production forecasting with the GraphCast model offers valuable insights for the renewable energy sector, enabling better planning and decision-making based on expected solar energy production. By providing accurate forecasts of the actual solar output, stakeholders can optimize resource allocation, grid management, and energy storage strategies.

Future work could explore further refinements to the model architecture, incorporating additional weather variables or leveraging advanced techniques. Additionally, the model could be extended to include other renewable energy sources, such as wind power, to provide a comprehensive forecasting solution for the energy sector.